\documentclass[10pt,twocolumn,letterpaper]{article}

\usepackage{cvpr}
\usepackage{times}
\usepackage{epsfig}
\usepackage{graphicx}
\usepackage{amsmath}
\usepackage{amssymb}
\usepackage{courier}
\usepackage{subfig}

\usepackage[pagebackref=true,breaklinks=true,letterpaper=true,colorlinks,bookmarks=false]{hyperref}

\cvprfinalcopy 

\def\cvprPaperID{898} 

\ifcvprfinal\fi
\begin{document}

\title{PASCAL Boundaries: A Class-Agnostic Semantic Boundary Dataset}

\author{Vittal Premachandran \qquad Boyan Bonev \qquad Alan L. Yuille\\
University of California, Los Angeles,\\
{\tt\small \{vittalp@, bonev@, yuille@stat.\}ucla.edu}
}

\maketitle

\begin{abstract}
   In this paper, we address the \emph{boundary detection} task motivated by the ambiguities in current definition of edge detection. To this end, we generate a large database consisting of more than 10k images (which is 20$\times$ bigger than existing edge detection databases) along with ground truth boundaries between 459 semantic classes including both foreground objects and different types of background, and call it the \emph{PASCAL Boundaries} dataset, which will be released to the community. In addition, we propose a novel deep network-based multi-scale semantic boundary detector and name it Multi-scale Deep Semantic Boundary Detector (M-DSBD). We provide baselines using models that were trained on edge detection and show that they transfer reasonably to the task of boundary detection. Finally, we point to various important research problems that this dataset can be used for.
\end{abstract}

\section{Introduction}

Edge detection has been a fundamental problem in computer vision since the 1970's \cite{fram1975quantitative}. Detecting edges is beneficial for many vision tasks, for example, object detection \cite{zitnick2014edge}, image segmentation \cite{arbelaez2011contour}, neural circuit reconstruction from brain images \cite{ciresan2012deep}, and autonomous navigation, among others. This problem is under active research and potential solutions include local filtering-based approaches like the Canny edge detector \cite{canny1986computational} and the zero-crossing algorithm \cite{torre1986edge}, to pixel-level classification methods that use features obtained by careful manual design like \textit{gPb} \cite{arbelaez2011contour}, to patch-based clustering algorithms such as Structured Edges (SE) \cite{dollar2014fast}, to the more-recent deep learning based approaches such as the N4-network \cite{ciresan2012deep} or HED \cite{xie2015holistically}. 

\begin{figure}
\includegraphics[width=0.49\linewidth]{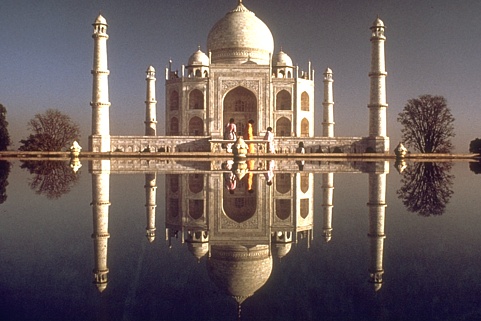}
\includegraphics[width=0.49\linewidth]{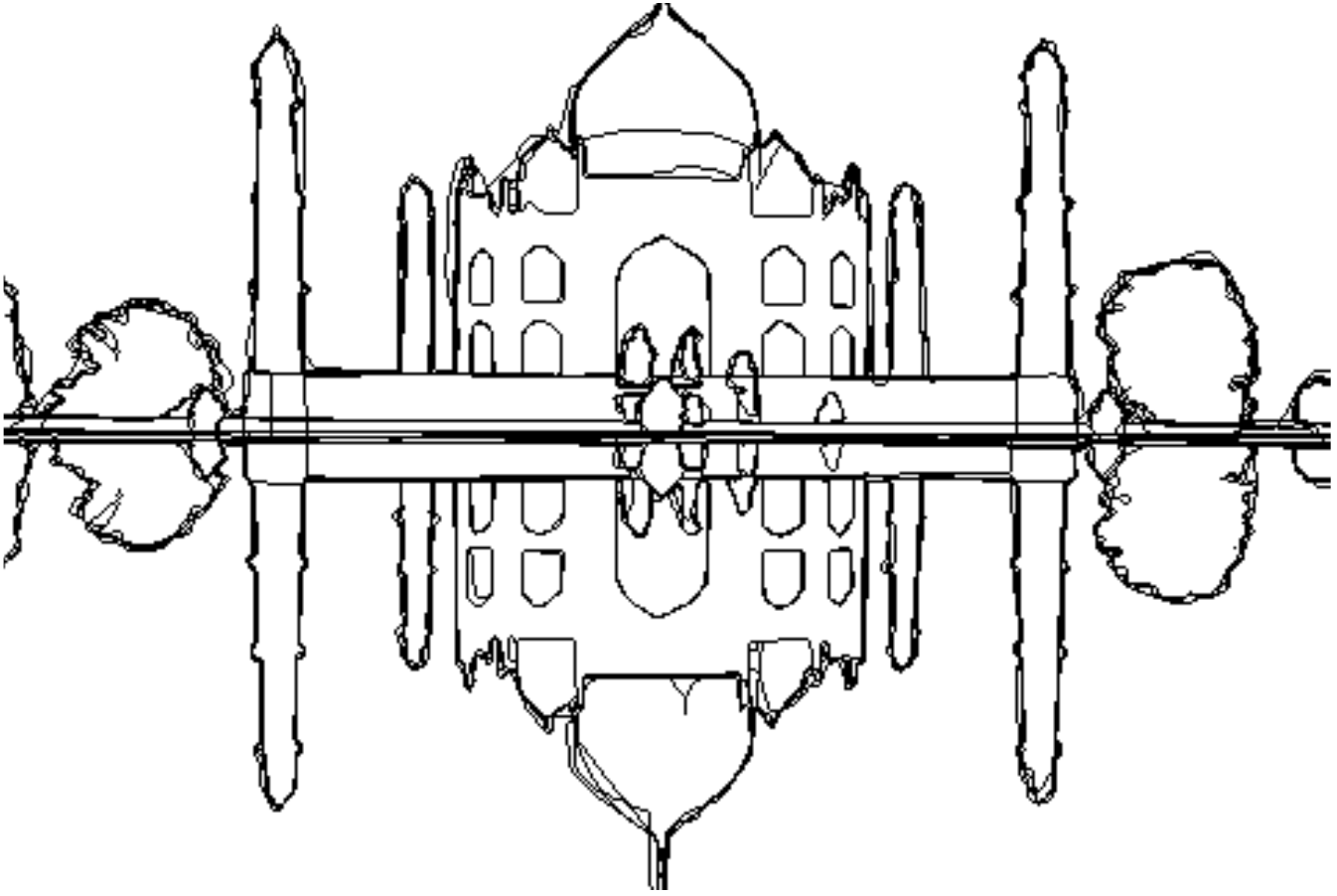}\\
\includegraphics[width=0.49\linewidth]{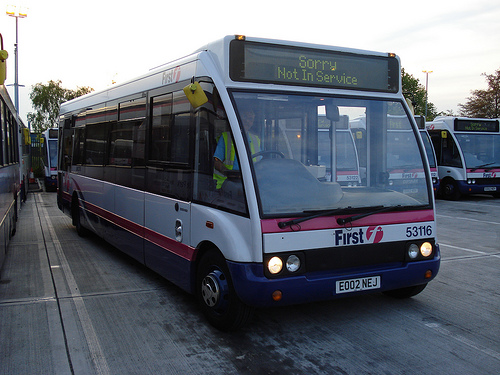}
\includegraphics[width=0.49\linewidth]{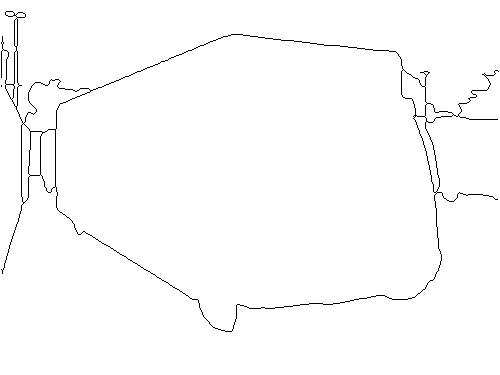}
\vspace{-2ex}
\caption{This figure shows the differences between edge annotations from the BSDS500 dataset (top row) and our class-agnostic object-level boundary annotations (bottom row). Our annotations are restricted to object outlines and background classes and generate boundaries around 459 semantic classes.}
\label{fig:teaser}
\vspace{-4.5ex}
\end{figure}
However, edge detection is an ambiguous task making it difficult to evaluate.
There is no clear answer to the question, `What is an edge?' An accepted definition of an edge is those sets of pixels with strong gradients.
Existing edge detection databases such as the BSDS300 \cite{martin2004learning} and BSDS500 \cite{arbelaez2011contour} were generated by asking the annotators to divide the image into multiple segments resulting in different annotators dividing the images into different segments. This lack of consistency arises because of the fact that edges can occur at different levels of granularity; i) just the exterior boundaries of objects, which divide the image into different object instances (car, road, tree, etc.), ii) interior boundaries dividing an object into its constituent parts (head, neck, torso, etc.), or iii) non-semantic contours emanating from texture (stripes on a tiger) or artificial design (writings on clothes). Hou et al. \cite{hou2013boundary} discuss the ambiguities in these datasets in more detail. In addition to the ambiguity, the BSDS500 dataset has only 500 images and cannot be considered as a large database. This motivates us to construct a new, large, class-agnostic semantic boundary dataset, that is not only large in comparison to BSDS500, but is also without the ambiguity of edge detection.

In this work, we wish to eliminate this ambiguity by restricting ourselves to the coarsest level of granularity, i.e., semantic instance-level object boundaries. Further extensions of our work is possible by introducing edges of other levels of granularity. Figure \ref{fig:teaser} shows the differences between the BSDS500 annotations (top row) and our annotations (bottom row).

This paper makes the following contributions: i) We define a precise task, namely, \textit{boundary detection}. To enable progress on this problem, we construct a large dataset with $\sim$10k images taken from the PASCAL VOC2010 challenge and provide boundary annotations between 459 semantic classes including both foreground objects and different types of background. These boundary annotations will be released publicly. The dataset generation process is described in more detail in Section \ref{sec:dataset}. ii) We propose a novel multi-scale deep network-based class-agnostic semantic boundary detector (M-DSBD) to solve the boundary detection task. This is described in detail in Section \ref{sec:multiscale-deepNetwork}.

The rest of the paper is organized as follows. In Section \ref{sec:exp}, we provide baselines on this new dataset using two well-performing edge detectors, i.e., Structured Edges (SE) \cite{dollar2014fast} and Holistically-Nested Edge detector (HED) \cite{xie2015holistically}. We also provide results obtained using M-DSBD, and its single-scale counterpart DSBD, using existing evaluation methodologies like the F-measure to enable fair comparisons with current and future approaches. Finally, we conclude the paper in Section \ref{sec:conclusion} by pointing to various future directions that this dataset paves way to, which would enable progress in many computer vision problems.

\section{Related Work}

One of the first databases for edge detection was the Sowerby database \cite{bowyer1998empirical}, which consisted of $100$ color images from the streets of Sowerby. At about the same time, there was the South Florida dataset \cite{bowyer1998empirical}, which was simpler than the Sowerby dataset, and consisted of $50$ gray scale images. These datasets enabled the identification of the fact that low-level filtering techniques such as the Canny edge detector \cite{canny1986computational} were limited in many ways. Moreover, the first real statistical techniques for edge detection \cite{konishi1999fundamental, konishi2003statistical} were developed and tested on these datasets.

The Sowerby dataset being too small motivated Martin et al. \cite{martin2001database} to start creating a public dataset of image segmentations.
A set of $300$ images from this dataset was then used in \cite{martin2004learning}, who cast the problem of edge detection as a per-pixel classification problem and evaluated their results using a precision-recall curve, which can be summarized by the now-standard F-Measure. This set of $300$ images later came to be known as the BSDS300 dataset. The BSDS300 dataset enabled the development of some notable edge detectors such as BEL \cite{dollar2006supervised}.

Recently, the BSDS300 dataset was extended to incorporate $200$ additional images \cite{arbelaez2011contour} and the new superset dataset was named as the BSDS500 dataset. The BSDS500 dataset has since been heavily worked upon producing significant efforts on edge detection algorithms such as the \textit{gPb}-edge detector \cite{arbelaez2011contour}, Sketch Tokens \cite{lim2013sketch}, SE \cite{dollar2014fast} and, the now state-of-the-art method, HED \cite{xie2015holistically}. Over the last couple of years, many deep network-based edge detection methods, such as the N4-network \cite{ciresan2012deep}, Deep Edge \cite{bertasius2014deepedge}, Deep Contour \cite{shendeepcontour} and HED \cite{xie2015holistically}, have shown significant improvements in terms of the F-Measure on the BSDS500 dataset. While these algorithmic improvements are welcome, we feel that further jumps in performance will be limited by the size of the BSDS500 dataset.

The issue with regards to the scale of the BSDS500 dataset and its ambiguity was addressed to a certain extent (though not consciously) by Hariharan et al.~\cite{hariharan2011semantic}. They built an instance-level segmentation dataset for the $20$ PASCAL object categories in about $10,000$ images corresponding to the trainval set of the PASCAL VOC2010 challenge. However, their aim in building this dataset was to tackle the problem of obtaining class-specific object boundaries, thus requiring them to train $\mathcal{O}(N)$ boundary detectors corresponding to $N$ object classes. Uijlings and Ferrari \cite{uijlings2015situational} go even more extreme by subdividing each object class into $K$ subclasses, which they call ``situations", thus requiring them to train $\mathcal{O}(NK)$ situational object boundary detectors. Clearly, neither of these approaches are scalable for large $N$ and $K$. Therefore, we adopt a class-agnostic boundary detection strategy by going back to building a strong monolithic boundary detector, thus requiring us to train a $\mathcal{O}(1)$ boundary detector.
Such an approach allows for the sharing of the computation involved in performing the mid-level task across various high-level vision tasks. Sharing of mid-level computations enables seamless scalability across multiple high-level visual tasks.

A similar dataset is the MS-COCO dataset \cite{mscoco}, which contains instance-level masks for 80 object categories and is much larger than the PASCAL dataset. However, this dataset also contains masks only for foreground objects. In comparison, we consider all objects in the image and an initial taxonomy established 459 semantic classes that includes both foreground objects and different types of backgrounds (e.g. sky, water, grass, etc.). Finally, Zhu et al. \cite{zhu2015semantic} recently proposed an amodal segmentation dataset, where they label the complete extents of an object (even if they are occluded) on the 500 images of the BSDS500 dataset. We restrict ourselves to the unoccluded parts of the object since labeling the occluded regions of objects would again lead to ambiguity in the task.

\section{Dataset Description}
\label{sec:dataset}
We propose a new dataset for boundary detection tasks and call it \textit{PASCAL Boundaries}. It is fundamentally different from the well-known BSDS500~\cite{arbelaez2011contour} dataset. BSDS500 allows annotators to divide the images into multiple segments without providing a precise definition of an edge. The annotations thus consists of edges from multiple levels of segmentation hierarchy.

PASCAL Boundaries follows a different approach. The labels are obtained using \emph{clear} instructions and \emph{unequivocal} criteria, so that there are no disagreements. We restrict the labels to be only those edges that separate one object instance from another object instance of the same class, or another object instance from a different class, or, from a background type. This ensures that our boundaries are consistent since the visible extent of a class is well-defined.

The annotators were asked to label all the pixels belonging to the same object (without sub-pixel precision). This annotation produced the PASCAL Context~\cite{mottaghi_cvpr14} annotated dataset, which uses the images of the PASCAL VOC2010 challenge (10,103 \texttt{trainval} images). To minimize human errors, the images were reviewed two times by different subjects. The boundary annotations in the proposed PASCAL Boundaries dataset are obtained by an automatic post-processing of the PASCAL Context region-based annotations. The boundaries are localized exactly between pixels having different category, or instance, labels. Ideally, they would be 0 width: right between the objects. In practice, we label the boundaries of both objects, which produces two pixel wide boundary annotations. This can be useful for some setups, but in our experiments we thinned them using morphological operations to have boundaries of one pixel width. We do not use sub-pixel precision in our annotations because we found that annotating at such levels of precision would be beyond the abilities of human annotators. Rows 1 and 3 in Figure \ref{fig:qualitative} shows multiple examples of image-boundary pairs. Row 1 contains the original images, row 3 is the class-agnostic semantic boundary map that we obtain from the PASCAL Context annotations (shown in row 2 of Figure \ref{fig:qualitative}).


Thus, PASCAL Boundaries is the first dataset which comprehensively annotates unoccluded image boundaries with an unequivocal criterion. Many of the images in this dataset are non-iconic, which means they contain multiple objects, not necessarily biased towards ``photography" images (one salient object in the center with high contrast with respect to the background). Minimizing this kind of bias is beneficial for realistic computer vision applications. BSDS500, on the other hand, consists of images with a dominant foreground object and without much clutter in the images. We also emphasize that the number of images in the PASCAL Boundaries dataset ($\sim10$k) is much larger than in existing datasets. The increased scale of the dataset provides more variation in the boundary types and is beneficial for learning deep models. Moreover evaluations on $\sim5$k images ensures a stricter test than evaluations performed on just a couple hundred images.



\noindent\textbf{Dataset Statistics:} PASCAL Boundaries has images of 360$\times$496 pixels on average, from which an average of 1.45\% of pixels are annotated as boundaries. This percentage is slightly lower than the 1.81\% of pixels annotated as edges in BSDS500, on images of 321$\times$481 pixels size. This is understandable since the BSDS annotations consisted of edges from the interiors of objects. This number drops to 0.91\% if we consider only those pixels that were labeled by all the annotators annotating the image.


\noindent\textbf{Extensions: } Many extensions of this dataset are possible. It is easy to annotate junctions in the image, i.e., regions in the image where there is a confluence of more than two contiguous objects. In some types of junctions, for example, in the case of T-junctions, these boundary confluences could act as cues for occlusion.

\begin{figure}
 \includegraphics[trim={2.3cm 1.5cm 4.2cm 0.8cm},clip, width=\linewidth, height=3.5cm]{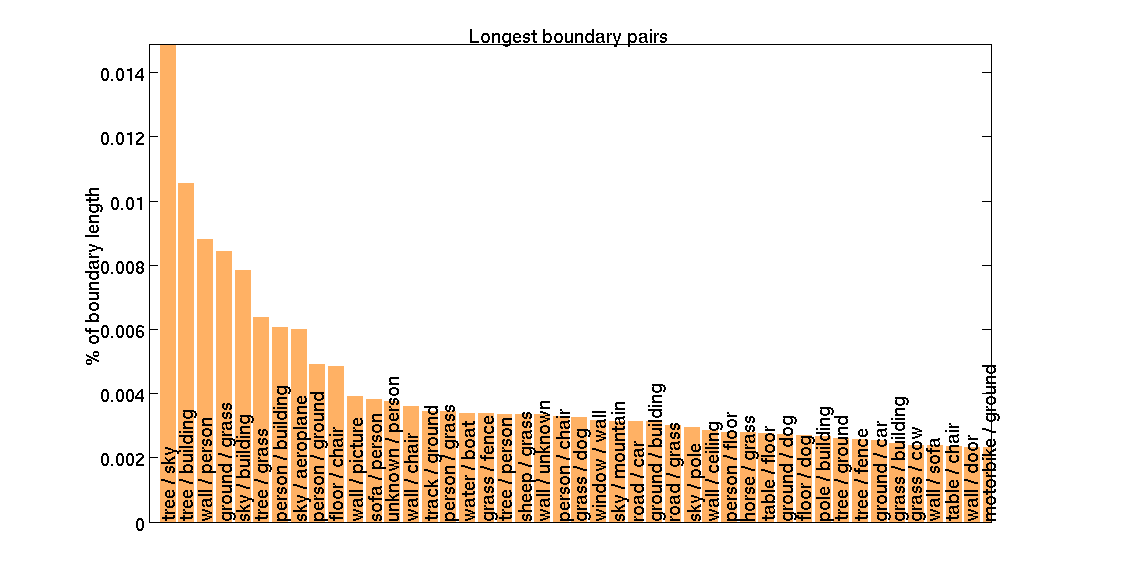}
 \caption{\label{fig:boundarypairs} Longest boundaries classified by the pairs of categories which the boundary separates. Only the top 45 out of 105,111 possible combinations are shown.}
 \vspace{-4ex}
\end{figure}

Another extension that we believe is useful is using the PASCAL Context class information in conjunction with the boundary information. In this way, the local appearances of boundaries can be analyzed and clustered based on pairs of classes on either side of a boundary. In Fig.~\ref{fig:boundarypairs} we show the most common shared boundaries, classified by pairs of categories, and sorted by boundary length. Note that the boundary length is influenced by the size of the regions in the image, not only by their number of instances.


\section{Multi-scale Deep Semantic Boundary Detector (M-DSBD)}
\label{sec:multiscale-deepNetwork}

\begin{figure}
\includegraphics[width=\linewidth, trim={7cm 2.5cm 6cm 4.5cm},clip]{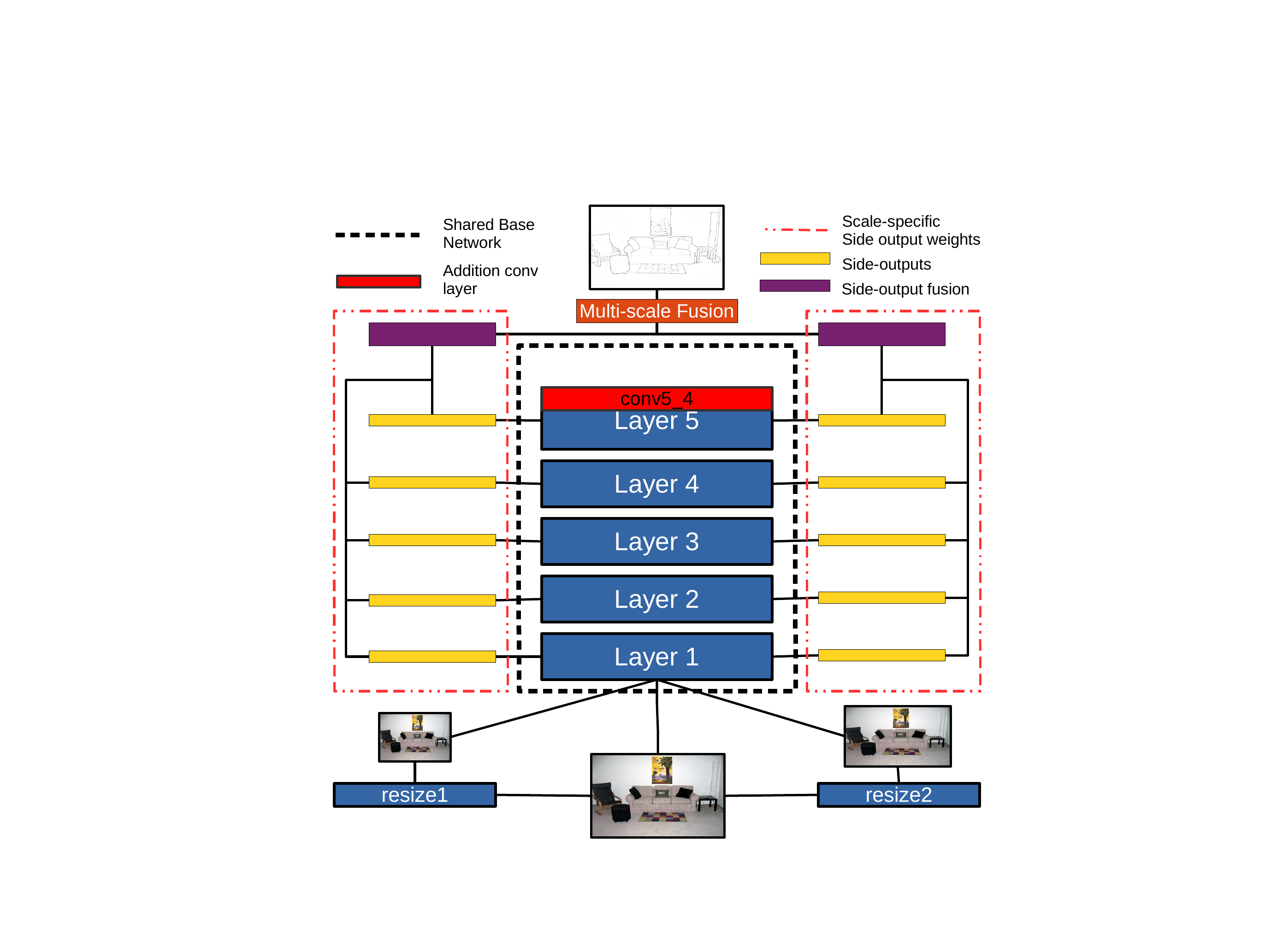}
\caption{This figure shows our multi-scale deep network architecture. The base network weights are shared across all scales. The figure only shows two side-output connections, while in practice, the multi-scale fusion layer fuses the predictions from three different scales.}
\label{fig:arch}
\vspace{-3ex}
\end{figure}

To complement the PASCAL Boundaries database, we propose a novel multi-scale deep network-based semantic boundary detector (M-DSBD). As an overview, our network takes an RGB image as input and outputs a prediction map that provides the confidence for the presence of a class-agnostic object boundary at each pixel location. To this end, we build upon the fully convolutional network (FCN) architecture \cite{long2014fully}.

Our network architecture is shown in Figure \ref{fig:arch}. M-DSBD works on multiple scales of input images, which is a common practice in many computer vision algorithms. Since the objects in our images occur at different scales, we try to provide invariance to it by \textit{explicitly} detecting object boundaries at various scales during both the training \textit{and} testing phases. Note that this is different from HED \cite{xie2015holistically}, where the authors use multi-scale only while training the network. Combining the predictions from multiple scales of the same image allows the deep network model to be scale-invariant, thus leading to a more robust boundary detector (also corroborated by the results from our experiments).

More formally, for a given image, $\mathbf{x}$, the network rescales it to multiple scales, $\mathcal{S} \in \{1,2,...,|\mathcal{S}|\}$, to produce an image pyramid, $\{\mathbf{x}^s\}_{s=1}^{|\mathcal{S}|}$. Our network acts on each rescaled image in this image pyramid, $\mathbf{x}^s$, and outputs a class-agnostic boundary map for each scale, $\hat{\mathbf{y}}^s(=\sigma(\hat{\mathbf{y}}^s_a))$. The final boundary prediction, $\hat{\mathbf{y}}$, involves taking a linear combination of the scale-specific boundary prediction activations, $\hat{\mathbf{y}}^s_a$,
\vspace{-2ex}
\begin{equation}
\hat{\mathbf{y}}(i) = \sigma(\sum_{s=1}^{|\mathcal{S}|} w^s_{scale} \hat{\mathbf{y}}^{s}_a(i)).
\vspace{-1ex}
\end{equation}
Here, $i$ is used to index the pixel locations in the image and $w^s_{scale}$ is the linear combination weight associated with scale $s$, which can be vectorized and written as $\mathbf{w}_{scale}$, and $\sigma(.)$ is used to denote the sigmoid function that maps the boundary prediction activations into the range $[0,1]$.

Each scale-specific boundary prediction, $\hat{\mathbf{y}}^s$, is obtained by passing the rescaled image, $\mathbf{x}^{s}$, though a series of convolutional layers, rectified linear unit (ReLU) layers, and max-pooling layers. We use $CNN(\mathbf{x}^{s}; \mathbf{W}_{base}, \mathbf{w}^{s}_{side})$ to denote the processing done on the rescaled image, $\mathbf{x}^{s}$, by a convolutional neural network parameterized by two sets of weights, $\mathbf{W}_{base}$ and $\mathbf{w}^{s}_{side}$, to produce the scale-specific boundary map,
\vspace{-2ex}
\begin{equation}
\hat{\mathbf{y}}^s = CNN(\mathbf{x}^{s}; \mathbf{W}_{base}, \mathbf{w}^{s}_{side}).
\end{equation}
Note that $\mathbf{W}_{base}$ is independent of the scale of the image and is shared across all image scales, and $\mathbf{w}^{s}_{side}$ denotes the scale-specific weights. We will explain both these weights in more detail, shortly.

Recently, various works have shown that a boost in performance is achievable by using features from the intermediate layers of the deep network \cite{long2014fully, hariharan2014hypercolumns, xie2015holistically}. M-DSBD also uses features from the intermediate layers of the base network, which we combine using a linear combination to produce a scale-specific boundary prediction map. Let, $\mathbf{f}^{(s,k)}(i) \in \mathbb{R}^{d_k}$ ($d_k$ is the number of convolutional filters in layer $k$) denote the feature vector at a spatial location, $i$, obtained from an intermediate layer, $k$, and, let the subset of the weights of the base network ($\mathbf{W}_{base}$) that are used to produce the features, $\mathbf{f}^{(.,k)}$, be denoted as $\mathbf{W}^{1:k}_{base}$. We fuse these features into a 1-channel feature map, $\mathbf{f}^{(s,k)}_{side}$, which can be extracted at the side of each intermediate layer, $k$, using a $1\times1$ convolution kernel, i.e.,
\vspace{-1ex}
\begin{equation}
\mathbf{f}^{(s,k)}_{side}(i) = {\mathbf{w}^{(s,k)}_{feat}}^{\top} \mathbf{f}^{(s,k)}(i)
\vspace{-1ex}
\end{equation}
where, $\mathbf{f}^{(s,k)}_{side}(i) \in \mathbb{R}$ is the 1-channel feature at the spatial location, $i$, and $\mathbf{w}^{(s,k)}_{feat}$ are the linear weights used to combine the intermediate layer features.

Due to the max-pooling at end of the intermediate layers, the spatial resolution of the side-output features, $\mathbf{f}^{(s,k)}_{side}$, will not be the same as the spatial resolution of the image, $\mathbf{x}^{s}$. So, we upsample the side-output features, using a deconvolution layer with an appropriately sized kernel, $\mathbf{w}^{(s,k)}_{up}$, before taking a linear combination of these side output features to produce the scale-specific boundary prediction activation,
\vspace{-1ex}
\begin{equation}
\hat{\mathbf{y}}^s_a(i) = \sum_{k=1}^{K}{w}^{(s,k)}_{fuse} \mathbf{f}^{(s,k)}_{(side, up)}(i).
\vspace{-1ex}
\end{equation}
Here, $\mathbf{f}^{(s,k)}_{(side, up)} = UP(\mathbf{f}^{(s,k)}_{side}; \mathbf{w}^{(s,k)}_{up})$ is the upsampled feature map, $\mathbf{w}^{(s,k)}_{up}$ are the weights corresponding to the interpolation kernel, and ${w}^{(s,k)}_{fuse} \in \mathbb{R}$ is the weight associated with the $k$-th layer side output for performing the linear fusion. We combine all linear fusion weights into a vector notation, $\mathbf{w}^{s}_{fuse} \in \mathbb{R}^K$, where K is the total number of layers in the deep network. We group all the side-output weights and denote the set as $\mathbf{w}^{s}_{side} = \{\mathbf{w}^{(s,k)}_{feat}\}_{k=1}^{K} \bigcup \{\mathbf{w}^{(s,k)}_{up}\}_{k=1}^{K} \bigcup \{\mathbf{w}^{s}_{fuse}\}$.

We initialize the base network weights, $\mathbf{W}_{base}$, from the five convolutional layers of the VGG16 network \cite{simonyan2014very}, which was pretrained on the ImageNet database. We encourage the reader to refer to \cite{simonyan2014very} for the architecture of the base network. From our experiments, we found that augmenting the VGG16 convolutional weights, with an additional convolutional layer (\texttt{conv5\_4}), improved the performance of the boundary detection task. Therefore, our base network architecture consists of the original convolutional layers from the VGG16 architecture and an additional convolutional layer, \texttt{conv5\_4}, which consists of 512 filters of size $3\times3$. The weights for this new \texttt{conv5\_4} layer were initialized randomly by drawing from a Gaussian distribution.

\subsection{Training Procedure}

We now describe the training procedure that was employed to train the weights in our deep network. As mentioned above, we build on the Fully Convolutional Network architecture, which allows us to backpropagate the gradients computed at each pixel location.

Our training set consists of the image-boundary label pairs, $\mathcal{D} = \{(\mathbf{x}_1,\mathbf{y}_1), (\mathbf{x}_2,\mathbf{y}_2),...,(\mathbf{x}_{|\mathcal{D}|},\mathbf{y}_{|\mathcal{D}|})\}$, where $\mathbf{x}_i$'s are the images and $\mathbf{y}_i$'s are the boundary labels. We employ batch-stochastic gradient descent to update the initialized weights. We make use of a layer-by-layer deep supervision \cite{lee2014deeply} to warm-start the training procedure. We greedily update the weights \cite{bengio2007greedy} in each layer by backpropagating the gradients from a side-output loss, $\Delta_k(\mathbf{y}, \hat{\mathbf{y}}^k)$, which is computed between the side output, $\hat{\mathbf{y}}^k(=\sigma(\mathbf{f}^{(s,k)}_{(side, up)}))$, obtained from the intermediate features out of layer $k$, and the ground truth boundary map, $\mathbf{y}$. The side-output loss is the sum of the weighted cross-entropy loss at each pixel location, i.e.,
\begin{equation}
\begin{aligned}
\Delta_k(\mathbf{y}, \hat{\mathbf{y}}^k) = - \beta \sum_{j \in \{i | \mathbf{y}(i) = 1\}} \log P(\hat{\mathbf{y}}^{k}(j)=1|\mathbf{x}; \mathbf{W}_{(\Delta,k)}) \\
- (1-\beta) \sum_{j \in \{i | \mathbf{y}(i) = 0\}} \log P(\hat{\mathbf{y}}^{k}(j)=0|\mathbf{x}; \mathbf{W}_{(\Delta,k)}),
\end{aligned}
\label{eq:side-loss}
\vspace{-1ex}
\end{equation}
where $\mathbf{W}_{(\Delta,k)} = \{\mathbf{W}^{1:k}_{base}\} \bigcup \{\mathbf{w}^{(s,k)}_{fuse}\} \bigcup \{\mathbf{w}^{(s,k)}_{up}\}$, and $\beta$ is the class-balancing weight. Class-balancing is needed because of the severe imbalance in between the number of boundary pixels and non-boundary pixels. We fixed $\beta = 0.9$, which we found to work well in our experiments.

The layer-by-layer deep supervision procedure uses a side-output loss, $\Delta_k(\mathbf{y}, \hat{\mathbf{y}}^k)$, to update only the weights corresponding to that layer. The weights of all other layers are not changed. For example, while backpropagating from $\Delta_k(\mathbf{y}, \hat{\mathbf{y}}^k)$, only the weights, $\{\mathbf{W}^{k}_{base}\} \bigcup \{\mathbf{w}^{(s,k)}_{fuse}\} \bigcup \{\mathbf{w}^{(s,k)}_{up}\}$ are updated; $\mathbf{W}^{k}_{base}$ corresponds to the weights in the $k$-th layer of the base network. The rest of the weights are untouched. We sequentially update the weights in each layer starting from layer $1$ and ending at layer $K$.

Once the weights have been fine-tuned using our greedy layer-by-layer update procedure, we switch off the side-output losses and finetune the network using a scale-specific boundary detection loss,
\vspace{-1ex}
\begin{equation}
\begin{aligned}
\Delta_s(\mathbf{y}, \hat{\mathbf{y}}^s) = - \beta \sum_{j \in \{i | \mathbf{y}(i) = 1\}} \log P(\hat{\mathbf{y}}^{s}(j)=1|\mathbf{x}; \mathbf{W}_{(\Delta,s)})\\
- (1-\beta) \sum_{j \in \{i | \mathbf{y}(i) = 0\}} \log P(\hat{\mathbf{y}}^{s}(j)=0|\mathbf{x}; \mathbf{W}_{(\Delta,s)}),
\end{aligned}
\label{eq:scale-scecific-loss}
\vspace{-1ex}
\end{equation}
where $\mathbf{W}_{(\Delta,s)} = \{\mathbf{W}_{base}\} \bigcup \{\mathbf{w}^{s}_{side}\}$. This is different from the training procedure in \cite{xie2015holistically}, where the authors employ deep supervision and force each side-output prediction to be a boundary map. We, on the other hand, only use deep supervision to warm-start the training procedure and switch off the gradients from the side-output loss while updating the fusion weights. In other words, we \textit{do not} enforce each side output to correspond to a boundary prediction map, but use these side outputs as features for the scale-specific boundary map. Enforcing each side output to be a boundary predictor of its own right prevents the fusion layer from providing the best performance. Allowing the side outputs to only act as features for the fusion layer, by switching off the gradients from the side-output loss, enables a layer's features to be complementary to other layers' features, thus permitting the fusion weights to extract the best possible performance.

All that is left is to learn the optimal weights to fuse the various scale-specific predictions. To this end, we define the final boundary detection loss, $\Delta_b(\mathbf{y}, \hat{\mathbf{y}})$ as,
\vspace{-1ex}
\begin{equation}
\begin{aligned}
\Delta_b(\mathbf{y}, \hat{\mathbf{y}}) = - \beta \sum_{j \in \{i | \mathbf{y}(i) = 1\}} \log P(\hat{\mathbf{y}}(j)=1|\mathbf{x}; \mathbf{W}_{(\Delta,b)}) \\
- (1-\beta) \sum_{j \in \{i | \mathbf{y}(i) = 0\}} \log P(\hat{\mathbf{y}}(j)=0|\mathbf{x}; \mathbf{W}_{(\Delta,b)}),
\end{aligned}
\label{eq:boundary-loss}
\vspace{-1ex}
\end{equation}
where $\mathbf{W}_{(\Delta,b)} = \{\mathbf{W}_{base}\} \bigcup \{\mathbf{w}^{s}_{side}\}_{s=1}^{|\mathcal{S}|} \bigcup \{\mathbf{w}_{scale}\}$. In this final stage of learning, we switched off the gradients from the side-output losses and the scale-specific losses, and backpropagated the gradients only from the boundary detection loss. Moreover, the base network weights $\mathbf{W}_{base}$ were not updated during this final stage, and only the side-output weights, $\{\mathbf{w}^{s}_{side}\}_{s=1}^{|\mathcal{S}|}$, and the scale-fusion weights, $\mathbf{w}_{scale}$, were updated.
\vspace{-2ex}
\section{Experiments}
\label{sec:exp}
\vspace{-1ex}
We predominantly experimented on the newly collected PASCAL Boundaries dataset that was described in Section \ref{sec:dataset}. This database consists of images from the \texttt{trainval} set of the PASCAL VOC2010 challenge. There are a total of 10,103 images that have been labeled. We train our deep network on the \texttt{train} set of the dataset and test on the \texttt{test} set. Note that since we label only the images from the \texttt{trainval} set of the PASCAL VOC2010 challenge, the \texttt{test} set of the PASCAL Boundaries dataset corresponds to the \texttt{val} set of the PASCAL VOC2010 challenge.

\noindent\textbf{Implementation details: }We used the publicly available FCN code \cite{long2014fully}, which is built on top of the Caffe framework to train our deep network. We modified the \texttt{sigmoid\_cross\_entropy\_loss\_layer} to compute the weighted cross entropy loss. In addition, we provide functionalities within the Caffe framework that resizes (downsample and upsample) data blobs to arbitrary resize factors\footnote{Source code will be released for public use.}. Weight updates were performed using batch-SGD with a batch size of $5$ images. To enable batch training on a GPU, we resized all images from the \texttt{train} set to a standard resolution of $400\times400$. The learning rate was fixed to 1e-7, and weight decay was set to 0.0002. We did not augment our training data since the PASCAL Boundaries dataset has $\sim5000$ training images.

\noindent\textbf{Evaluation Criterion:} The standard evaluation criterion for evaluating edge detection algorithms is the F-score. We also use the same evaluation criterion for evaluating our boundary detector. In addition, we provide baselines on the new dataset using two other well-known edge detection algorithms; SE \cite{dollar2014fast} and HED \cite{xie2015holistically}. We use the helper evaluation functions provided in the SE Detection Toolbox \cite{dollar_edges_code} to obtain all the numbers we report in this paper.

\subsection{Transfer from Edge Detection}
\begin{figure}
\includegraphics[trim={0cm 0.2cm 0cm 0cm},clip,width=\linewidth]{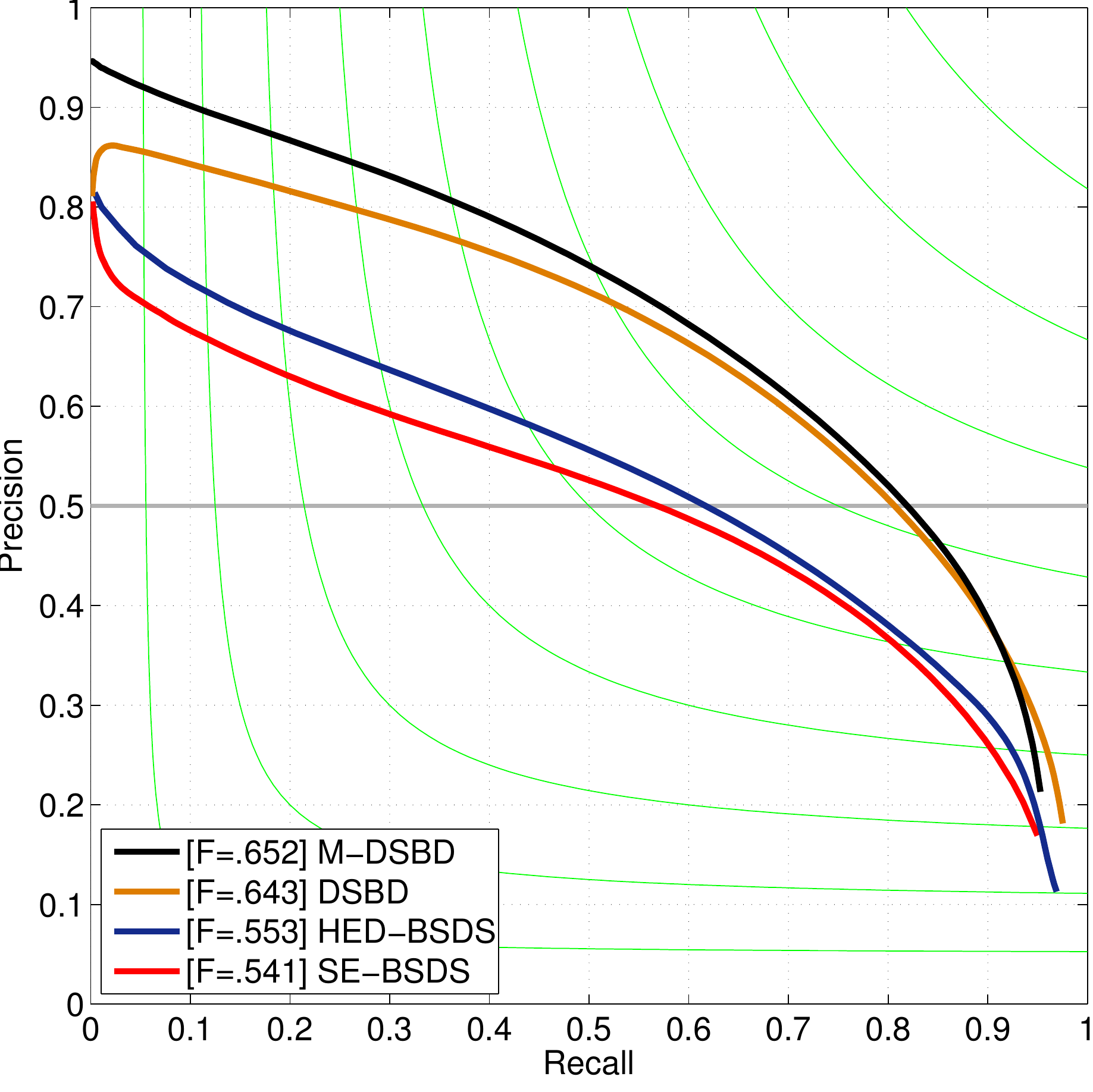}
\vspace{-4ex}
\caption{This plot shows the Precision/Recall curves on the PASCAL Boundaries dataset. The SE and HED curves were obtained using models trained on an edge detection task on the BSDS500 dataset. The results show that they transfers reasonably onto the boundary detection task. Results from M-DSBD shows that multi-scale processing of images produces better boundary maps.}
\label{fig:pasal-plot}
\vspace{-3ex}
\end{figure}

We tested the baseline edge detection methods, SE \cite{dollar2014fast} and HED \cite{xie2015holistically}, on the 5105 images present in the \texttt{test} set of the PASCAL Boundaries dataset, and Figure \ref{fig:pasal-plot} shows the precision/recall curves. A more detailed, and exhaustive comparison is provided in Table \ref{tab:pascal}. SE and HED models were trained on the BSDS dataset and were released by the respective authors. To make this explicit, we call them SE-BSDS and HED-BSDS, respectively.

We see that both SE-BSDS and HED-BSDS transfer reasonably on to the PASCAL Boundaries dataset; SE-BSDS achieves an F-score of 0.541, while HED-BSDS achieves an F-score of 0.553. The ranking order of SE's and HED's performance when tested on the BSDS500 dataset also transfers over when tested on the PASCAL Boundaries dataset. This shows that BSDS500 edges are not entirely different from our definition of segment boundaries. The BSDS500 boundaries constitute object-level boundaries, object part-level boundaries, and boundaries emanating texture. Our database, in comparison, deals only with object-level boundaries.

\noindent\textbf{Retraining HED on PASCAL Boundaries:} To provide a fair comparison, we tried training HED using their publicly-released training code. We retained all the parameters that were set by the authors. We only replaced the training set from the BSDS500's augmented training set (which HED uses) to PASCAL Boundaries' \texttt{train} set. To account for an increase in the complexity of the PASCAL Boundaries dataset (in comparison to the BSDS500 dataset), we trained HED for a total of 100k iterations (as opposed to the 10k terations that the authors report in \cite{xie2015holistically}). We snapshotted the model every 1000 iterations and used a validation set of $25$ images (randomly chosen from the \texttt{train} set) to select the best model. Surprisingly, we found the performance of the best model when tested on the PASCAL Boundaries' \texttt{test} set to be worse by several $\%$ points compared to the performance obtained using their released model (HED-BSDS)\footnote{We obtained an F-score of 0.36 when we trained HED using the authors' training code on the PASCAL Boundaries datset. Moreover, we also noticed a drop in performance when we tried replicating HED's results on the BSDS500 dataset.}. We believe that the optimal parameters set by the authors of HED to train on the BSDS500 dataset might not be the optimal parameters to train on the PASCAL Boundaries dataset. We did not experiment with different parameter settings.

\begin{figure*}
\includegraphics[width=0.19\linewidth]{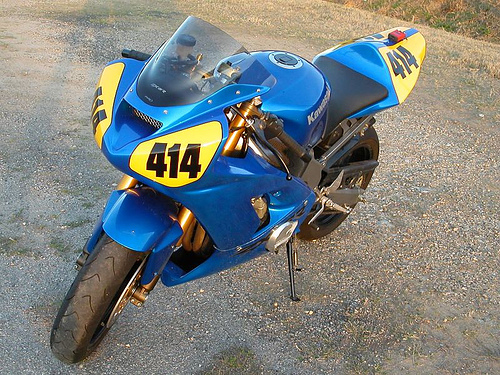}
\includegraphics[width=0.19\linewidth]{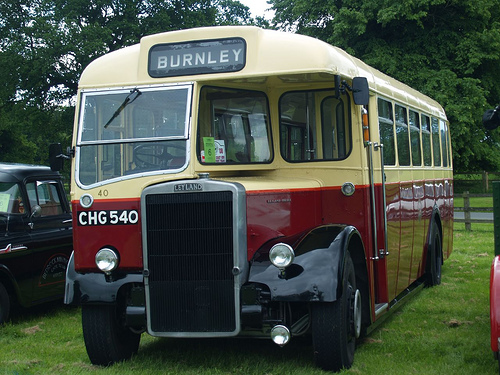}
\includegraphics[width=0.095\linewidth]{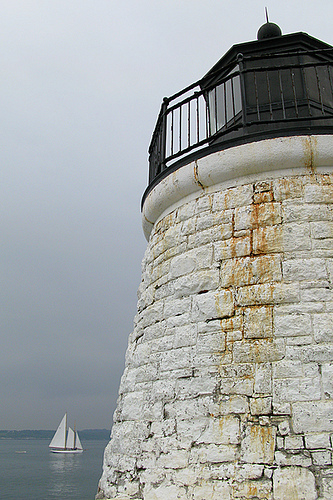}
\includegraphics[width=0.215\linewidth]{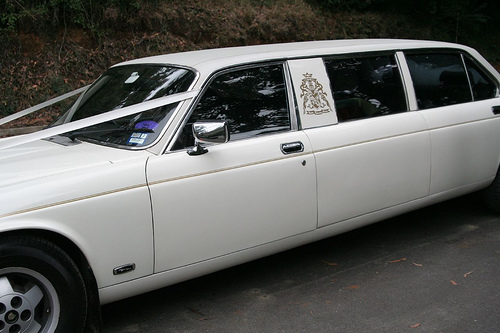}
\includegraphics[width=0.19\linewidth]{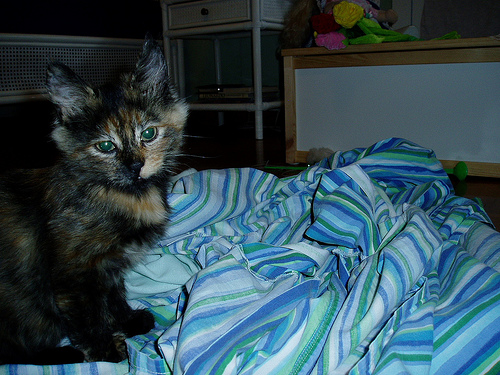}
\includegraphics[width=0.088\linewidth]{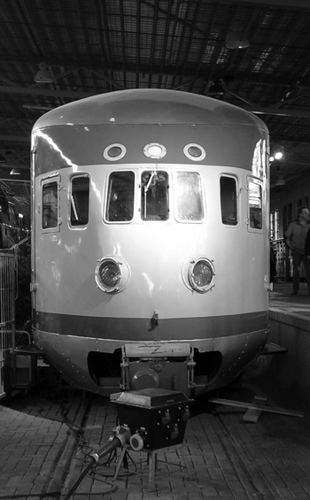}\\
\includegraphics[width=0.19\linewidth]{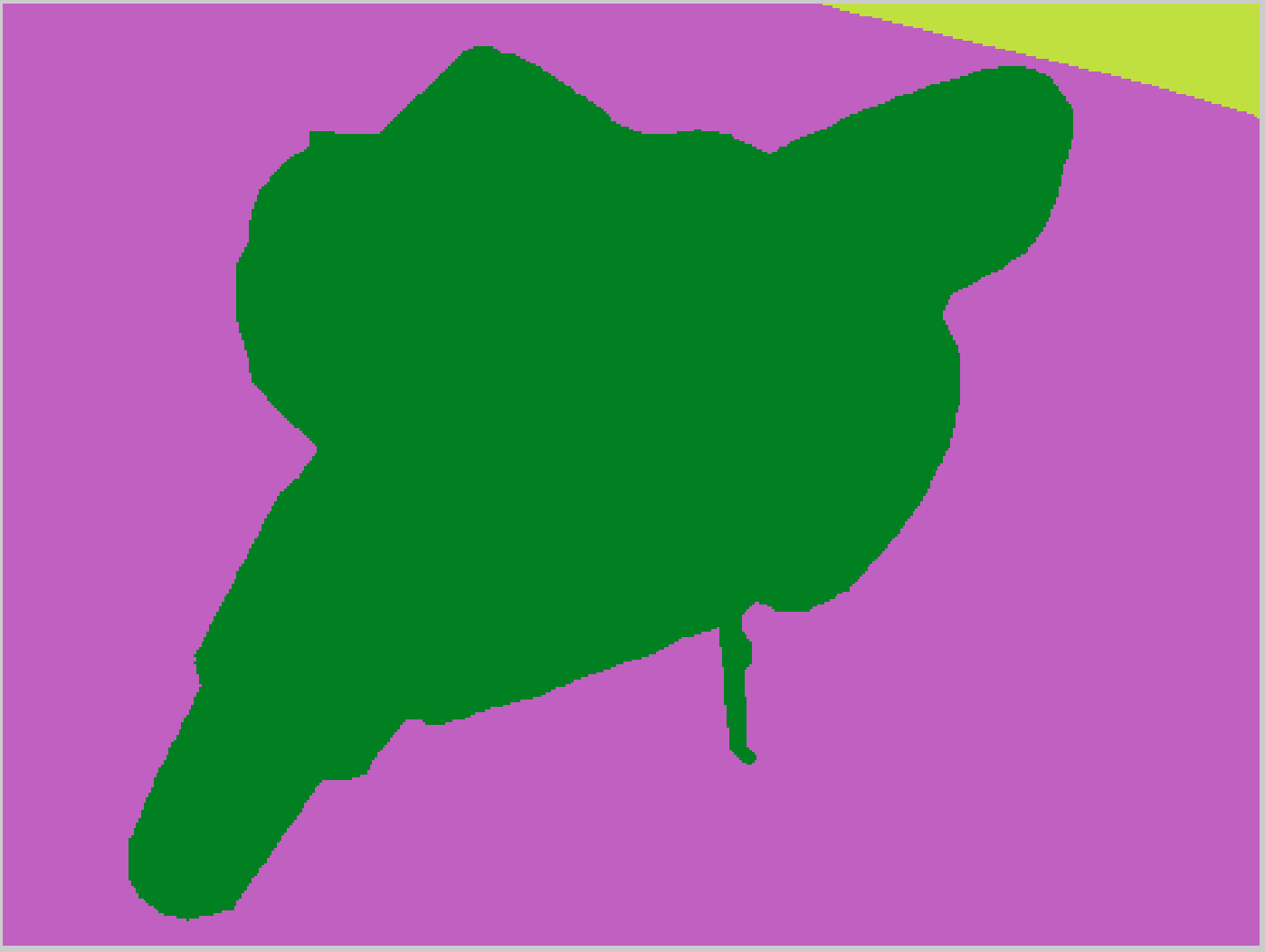}
\includegraphics[width=0.19\linewidth]{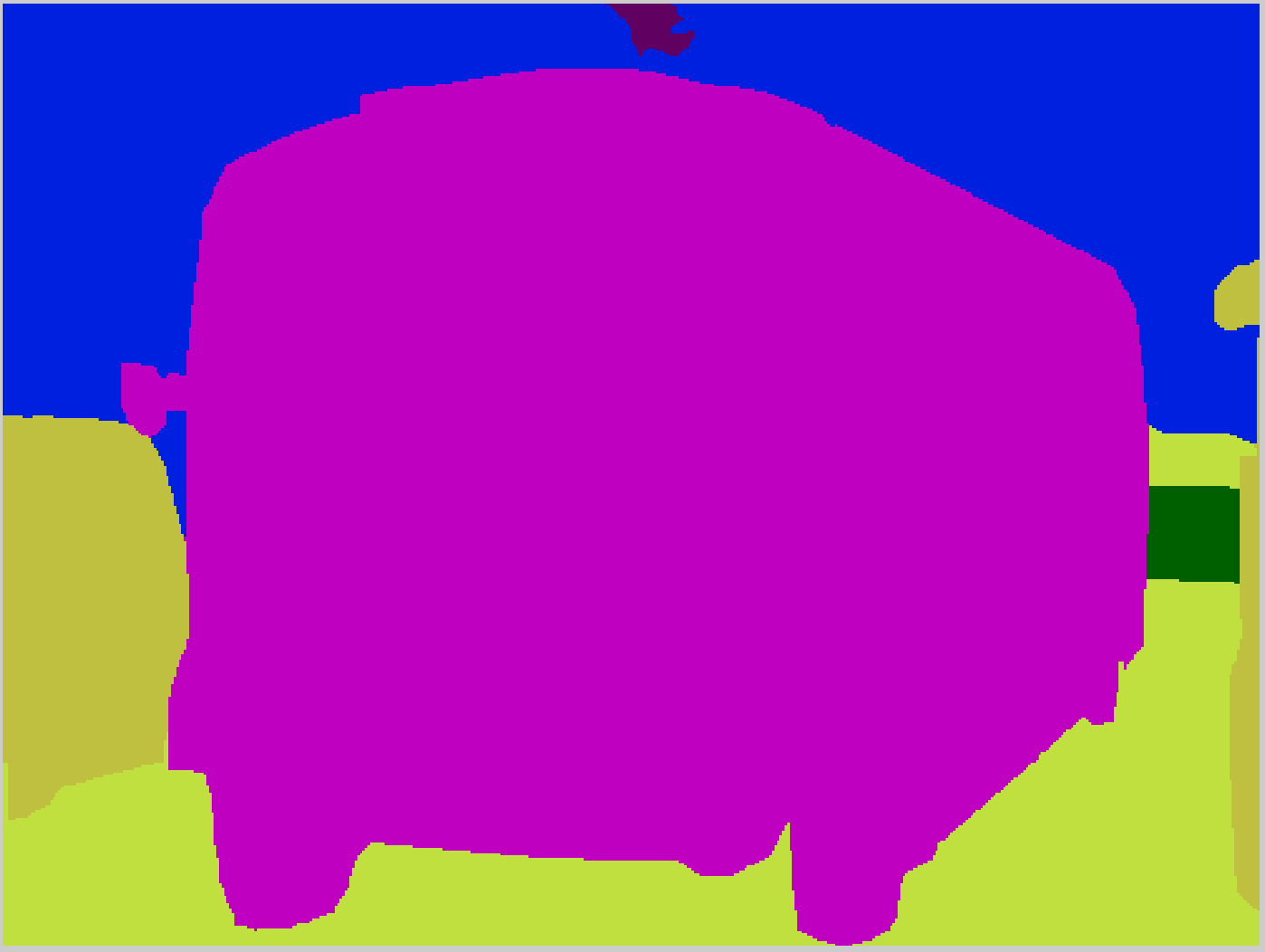}
\includegraphics[width=0.095\linewidth]{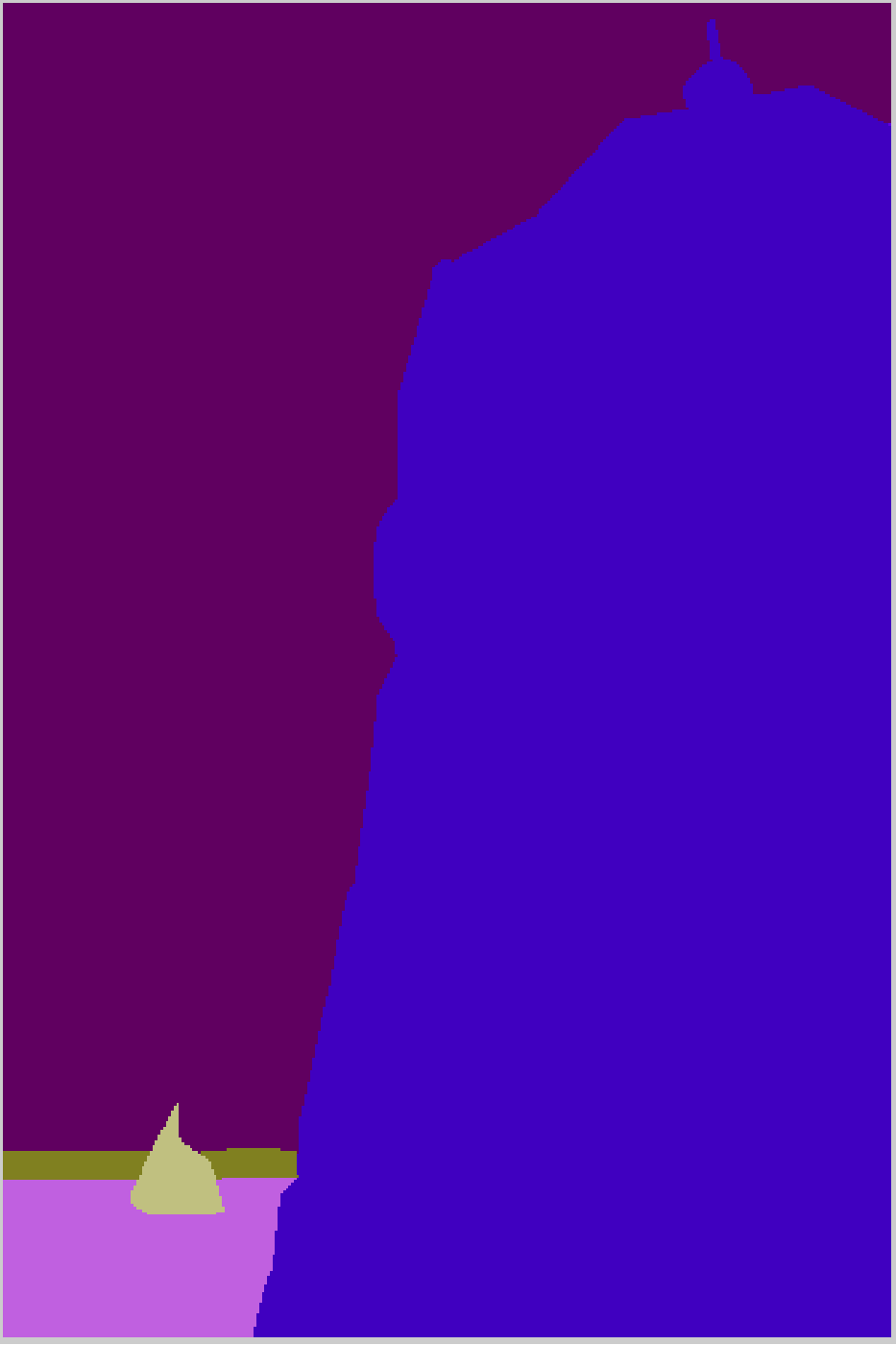}
\includegraphics[width=0.215\linewidth]{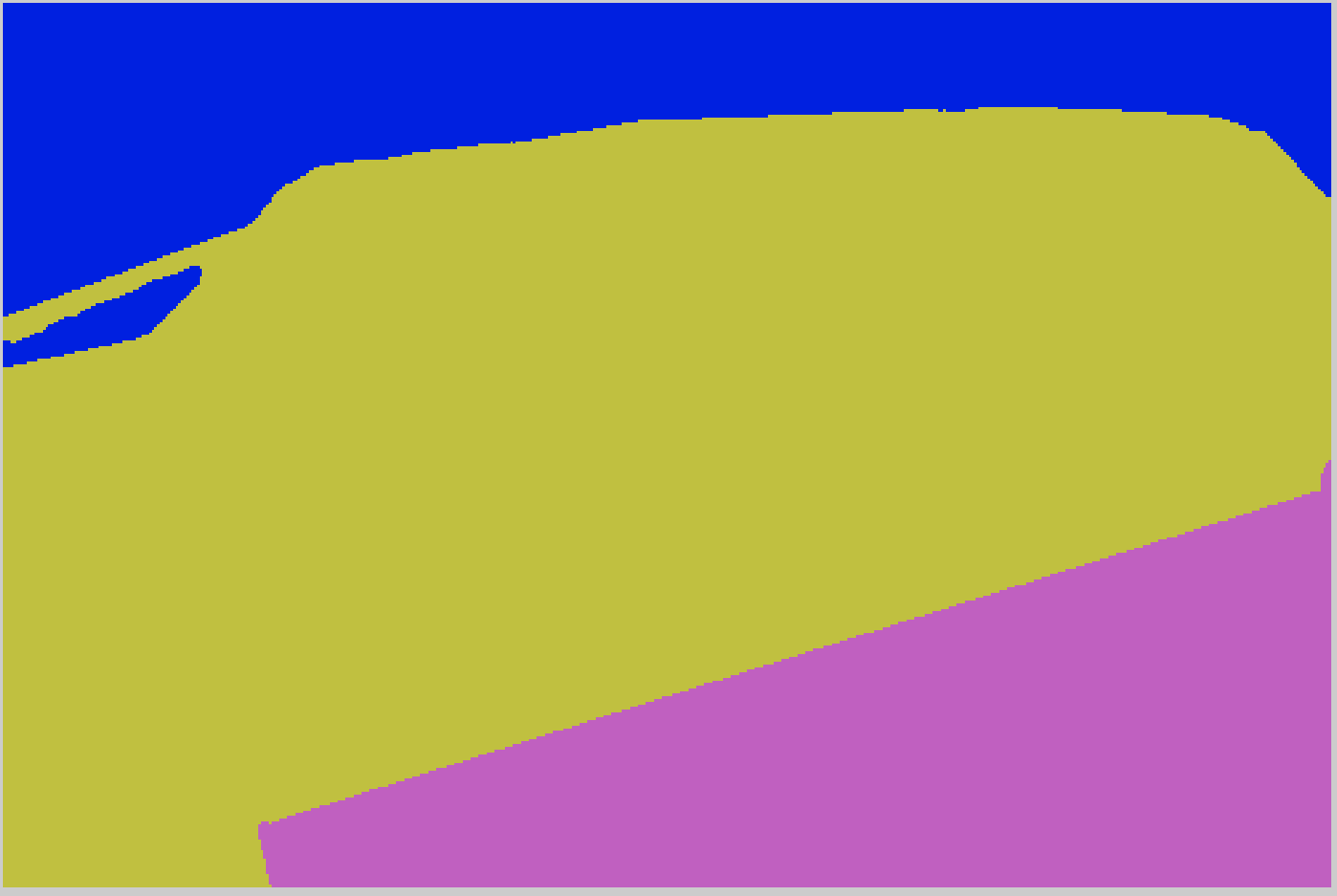}
\includegraphics[width=0.19\linewidth]{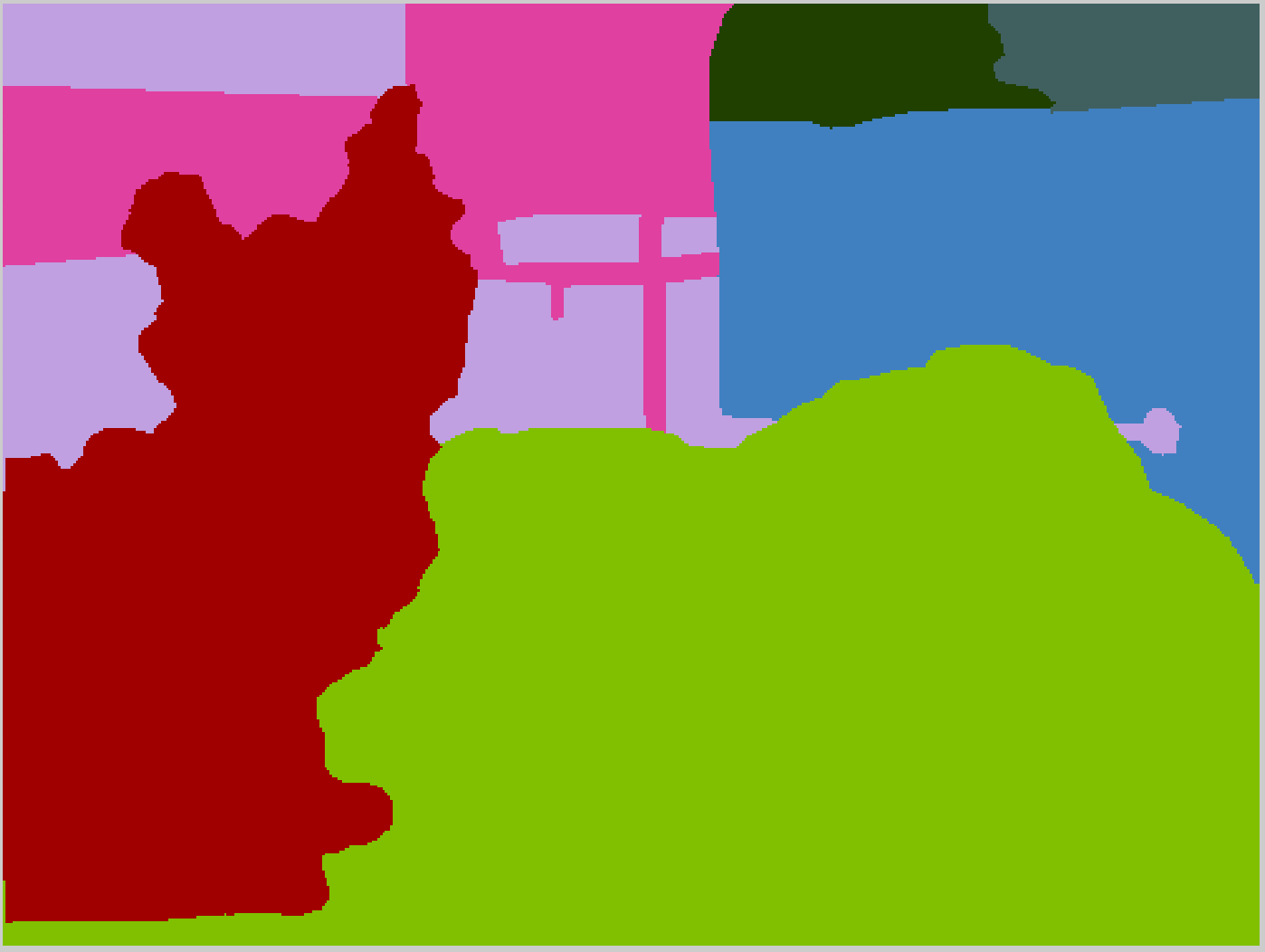}
\includegraphics[width=0.088\linewidth]{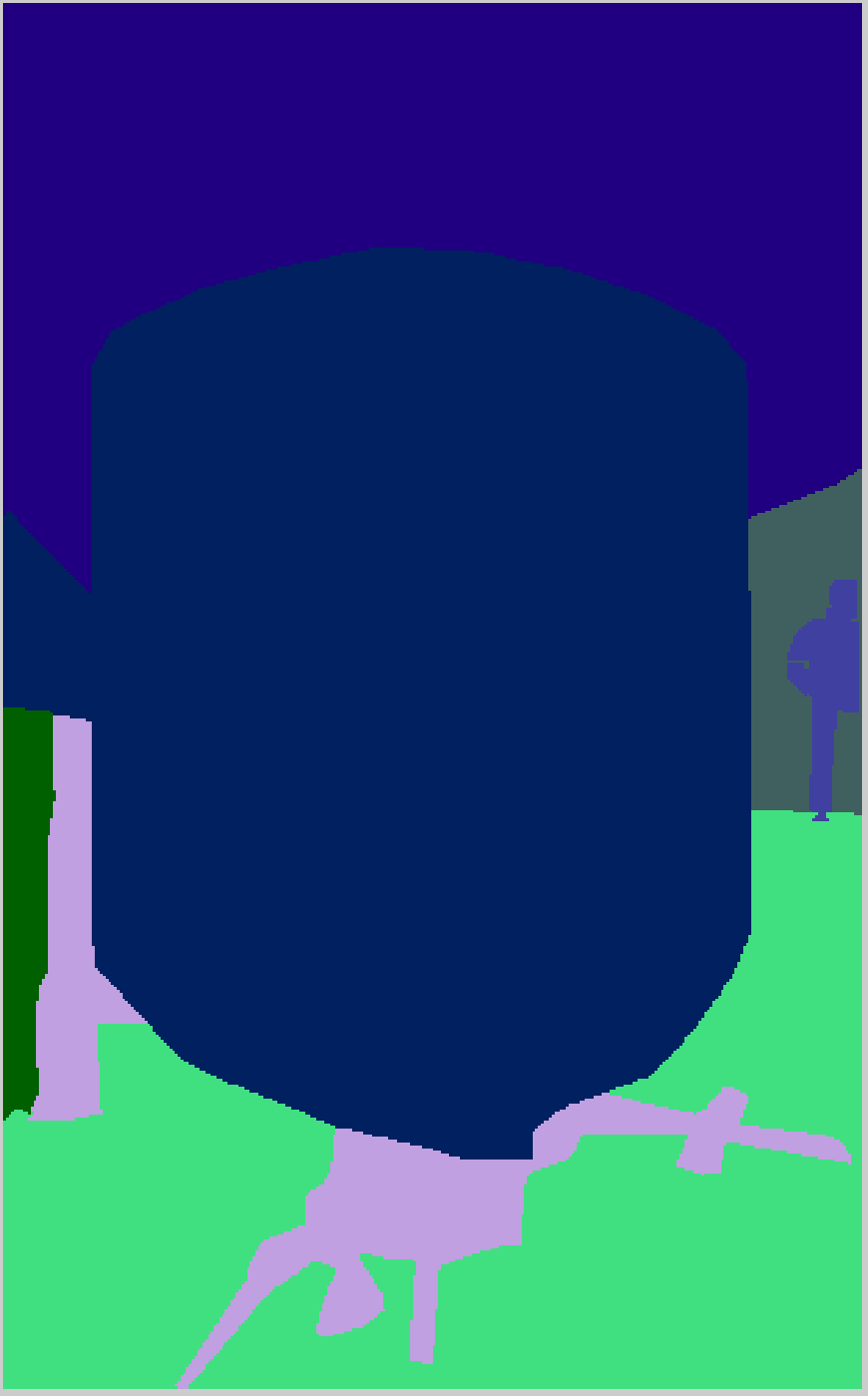}\\
\includegraphics[width=0.19\linewidth]{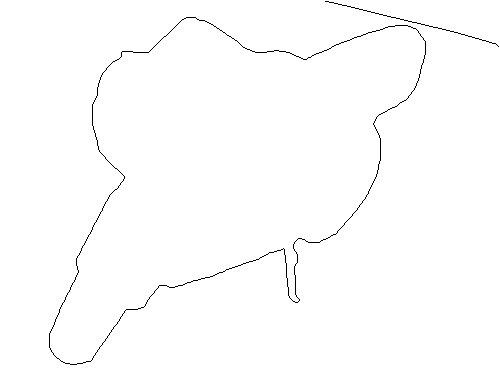}
\includegraphics[width=0.19\linewidth]{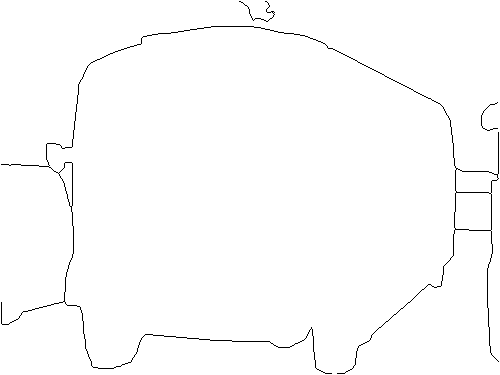}
\includegraphics[width=0.095\linewidth]{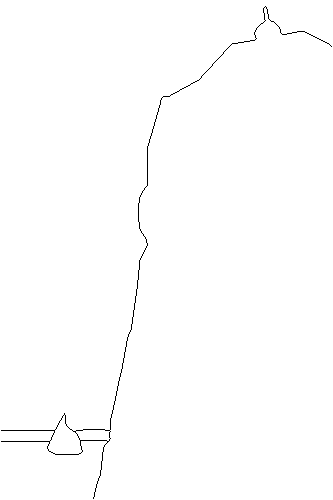}
\includegraphics[width=0.215\linewidth]{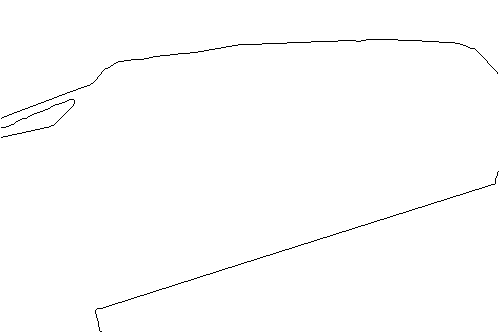}
\includegraphics[width=0.19\linewidth]{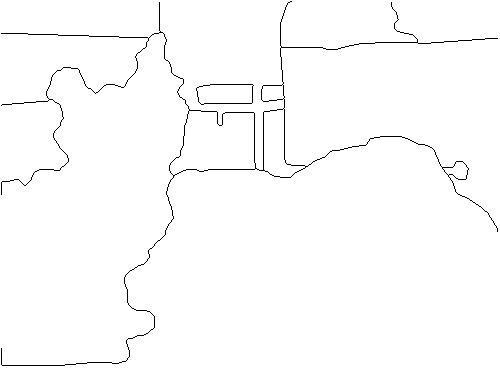}
\includegraphics[width=0.088\linewidth]{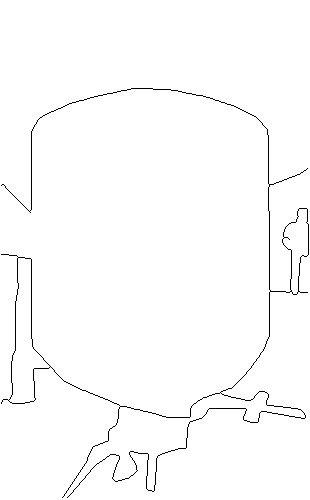}\\
\includegraphics[width=0.19\linewidth]{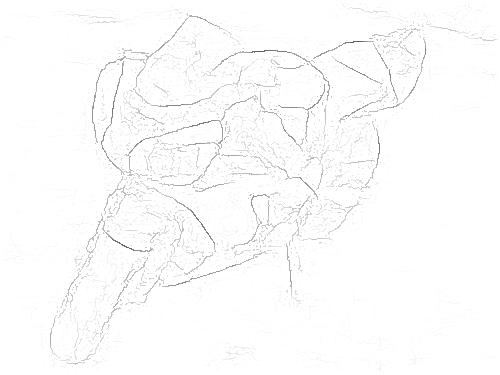}
\includegraphics[width=0.19\linewidth]{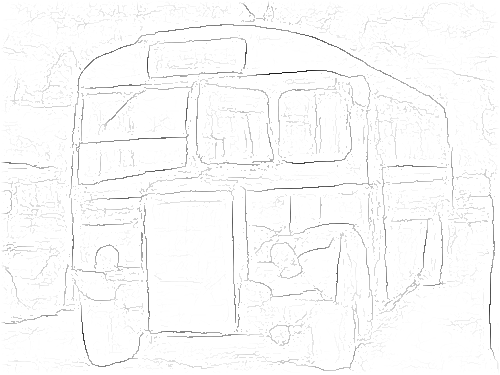}
\includegraphics[width=0.095\linewidth]{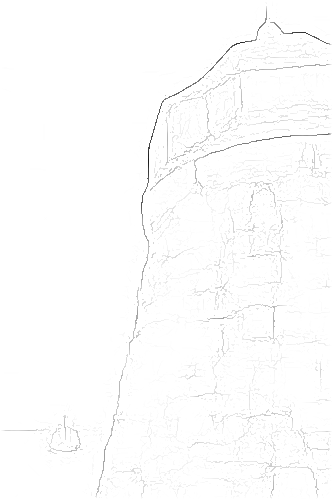}
\includegraphics[width=0.215\linewidth]{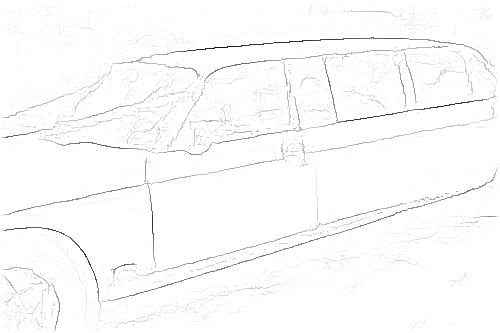}
\includegraphics[width=0.19\linewidth]{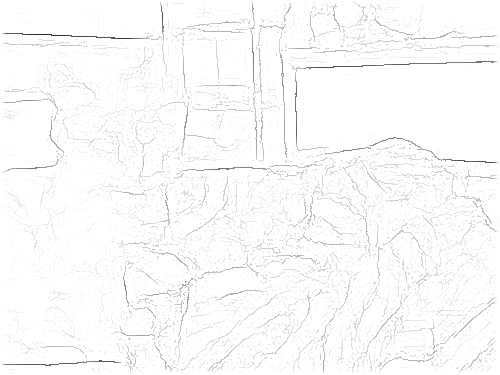}
\includegraphics[width=0.088\linewidth]{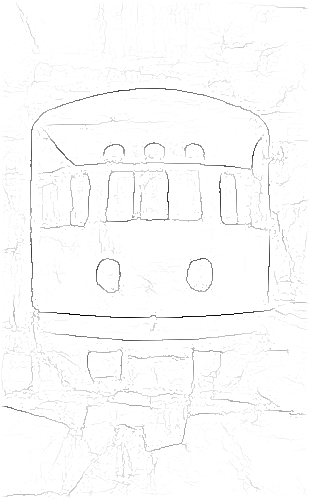}\\
\includegraphics[width=0.19\linewidth]{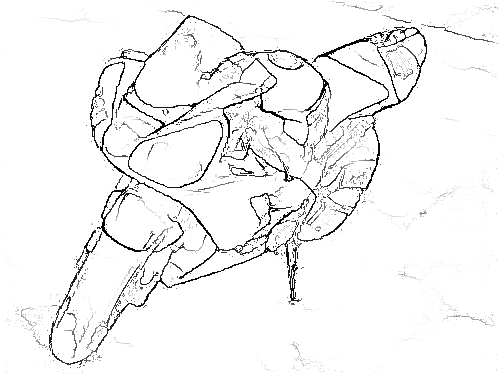}
\includegraphics[width=0.19\linewidth]{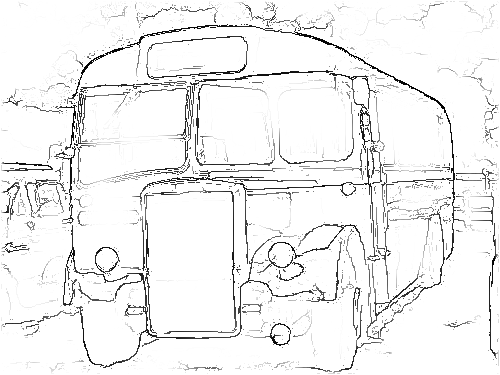}
\includegraphics[width=0.095\linewidth]{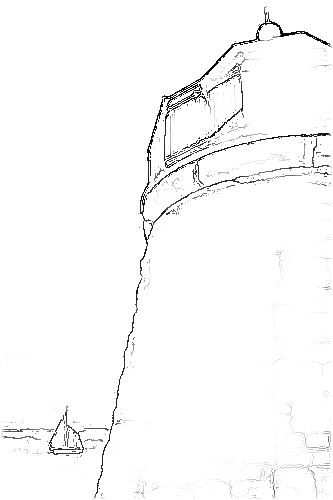}
\includegraphics[width=0.215\linewidth]{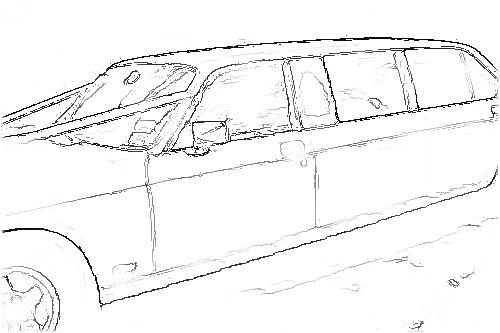}
\includegraphics[width=0.19\linewidth]{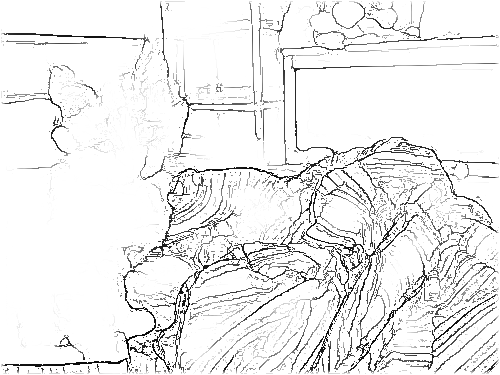}
\includegraphics[width=0.088\linewidth]{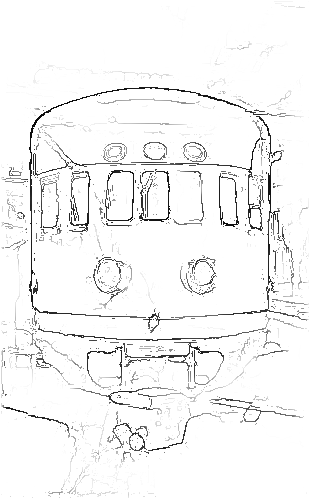}\\
\includegraphics[width=0.19\linewidth]{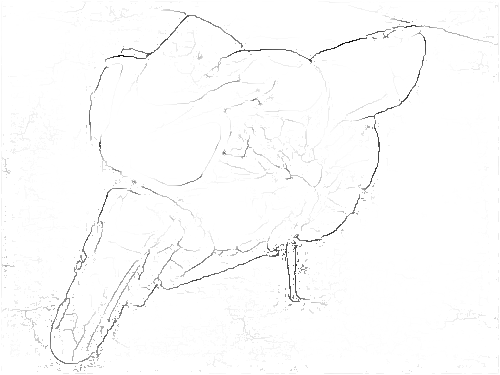}
\includegraphics[width=0.19\linewidth]{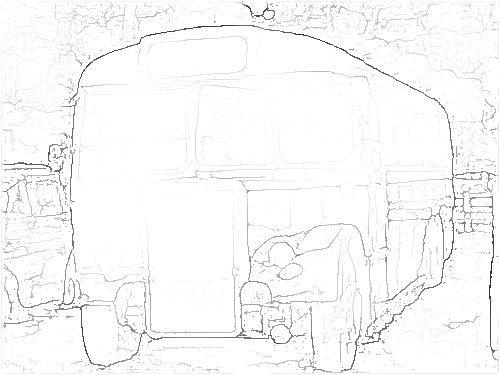}
\includegraphics[width=0.095\linewidth]{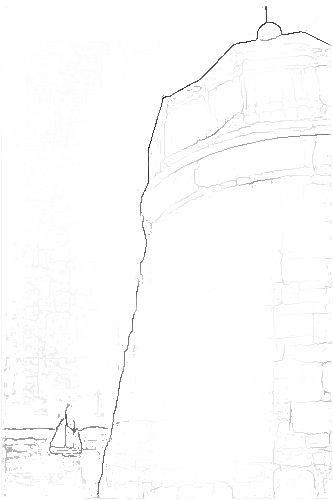}
\includegraphics[width=0.215\linewidth]{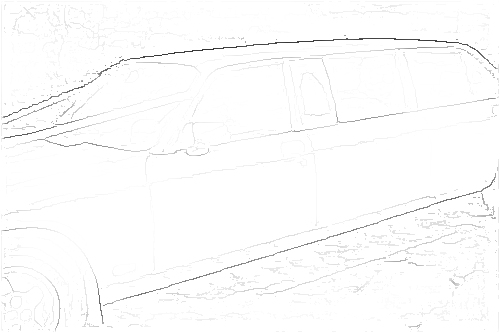}
\includegraphics[width=0.19\linewidth]{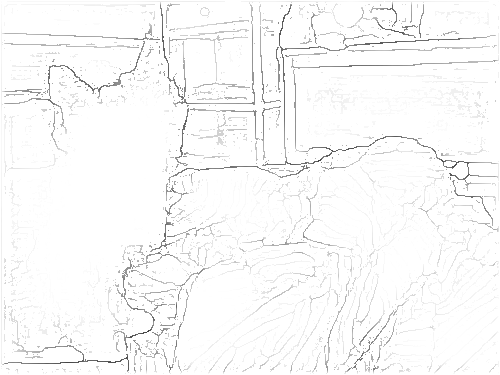}
\includegraphics[width=0.088\linewidth]{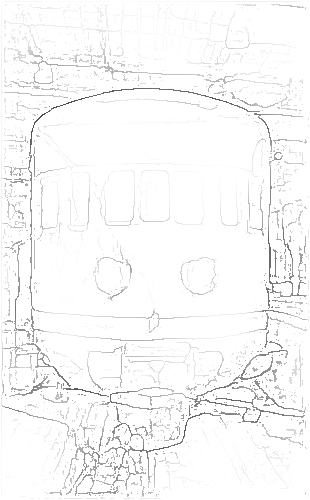}
\vspace{-2ex}
\caption{This figures shows some qualitative results. Row 1 shows example images, row 2 shows the respective per-pixel class annotations from the PASCAL Context dataset \cite{mottaghi_cvpr14}, which is used to generate the class-agnostic boundary maps of PASCAL Boundaries shown in row 3, rows 4 and 5 show results from SE \cite{dollar2014fast} and HED \cite{xie2015holistically}, respectively, and the final row shows the results from M-DSBD. Notice how M-DSBD is able to identify object-level boundaries and outputs far less number of internal edges. The edge detection techniques, on the other hand, detect edges across multiple levels of hierarchies.}
\label{fig:qualitative}
\vspace{-4ex}
\end{figure*}

\subsection{Accretion Study and (M-)DSBD Results}

We present our results in a step-by-step modular fashion to show the improvements that the respective components in our model provide.

\noindent\textbf{Training Strategy:} To test our training strategy, we replaced HED's training method with greedy layer-by-layer training strategy to warm-start the training process. We then used these updated weights as an initialization to train the HED architecture by backpropagating the gradients from the losses computed at each of the five side-output predictions and the fusion layer prediction, simultaneously, as was done in \cite{xie2015holistically}. This approach of training the HED architecture provided an improvement of $3\%$ over the results that were obtained while testing with the publicly-released pre-trained model; we were able to obtain an F-score of 0.59. Since this method uses the HED architecture but a different training strategy (i.e., the greedy layer-wise training), we use the term `HED-arch-greedy' to indicate this model. 

\noindent\textbf{More Convolutional Layers: }Since the PASCAL Boundaries dataset is more complex than the BSDS dataset, we experimented with adding more layers to the models so that it could capture the dataset's complexity. We began by adding an additional convolution layer, \texttt{conv5\_4}. We built the layer \texttt{conv5\_4} with $512$ filters, each with a kernel size of $3\times3$. We also added a ReLU layer to rectify the output of \texttt{conv5\_4}. This enhanced architecture was able to further improve the results by $3\%$ over the previous model by producing an F-score of 0.62 on the \texttt{test} set. We experimented with adding more layers to the network, but found that they did not improve the performance of the model. We use the term `HED-arch+\texttt{conv5\_4}-greedy' for this model.

\noindent\textbf{Switching deep supervision off:} An interesting outcome was observed when we used deep supervision just to warm-start the training process. Upon completion of the greedy layer-by-layer training process, we switched off the backpropagation of the gradient from the side-output losses (Eq. \ref{eq:side-loss}) and backpropagated only from the scale-specific boundary detection loss\footnote{Please note that the above experiment was done on a single scale. When we use the term ``scale-specific loss", the gradients were backpropagated from the loss computed using the original-sized images.} (Eq. \ref{eq:scale-scecific-loss}). Doing so, improved the performance of the model by another $2\%$. We call this version as the single scale Deep Semantic Boundary Detector (DSBD). We believe that the improvement in performance was achievable because we no longer force the side-output predictions to be boundary detectors of their own right, but use them as features for the fusion layer. That said, we do acknowledge the importance of deep supervision for warm-starting the training process.

\noindent\textbf{Multi-scale Boundary Detection:} Finally, we experimented with the M-DSBD architecture that was described in Section \ref{sec:multiscale-deepNetwork}. We used three scales, $\mathcal{S} = \{1, 0.8, 0.5\}$, for training and testing. The base network weights were not updated at this stage. Only the scale-specific side output weights, and the multi-scale fusion weights were updated during this final training procedure. The gradients were backpropagated from the boundary detection loss (Eq. \ref{eq:boundary-loss}). Our experiments supported our hypothesis that multi-scale processing would improve the task of boundary detection by providing a further improvement of $1\%$ on the \texttt{test} set of the PASCAL Boundaries dataset. Our model and training procedure produced a final F-score of 0.652, which is significantly more than the other baselines.

We tabulate all the numbers described above in Table \ref{tab:pascal}. `BSDS' is used to indicate that the model was trained on the BSDS500 dataset.
We also show some qualitative results in Fig. \ref{fig:qualitative}. Notice that our boundary detector is capable of identifying the semantic boundaries confidently and detects far less number of internal edges. On the other hand, the edge detectors identify edges across various levels of granularity (which they were trained to detect).


\begin{table}
\centering
\begin{tabular}{|c|c|c|c|}
\hline
Method & ODS & OIS & AP\\
\hline \hline
SE-BSDS \cite{dollar2014fast} & 0.541 & 0.570 & 0.486 \\
HED-BSDS \cite{xie2015holistically} & 0.553 & 0.585 & 0.518 \\
\hline
HED-arch-greedy & 0.59 & - & - \\
HED-arch+\texttt{conv5\_4}-greedy & 0.62 & - & - \\
DSBD & 0.643 & 0.663 & 0.650 \\
M-DSBD & \textbf{0.652} & \textbf{0.678} & \textbf{0.674} \\
\hline
\end{tabular}
\vspace{-2ex}
\caption{Results on the PASCAL Boundaries dataset. SE's and HED's results are from the models that were trained on the BSDS500. The results from M-DSBD shows that multi-scale does improve performance over single scale.}
\label{tab:pascal}
\vspace{-4ex}
\end{table}

\begin{figure}
\subfloat[Original Image]{\includegraphics[width=0.49\linewidth]{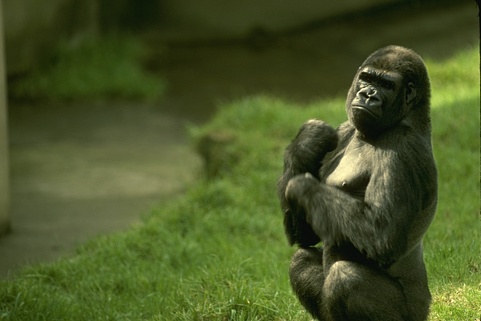}}
\subfloat[GT Edge Annotations]{\includegraphics[width=0.49\linewidth]{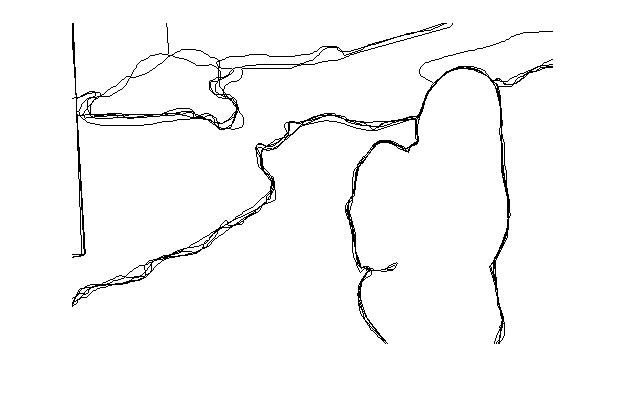}}\\
\subfloat[HED\cite{xie2015holistically}]{\includegraphics[width=0.49\linewidth]{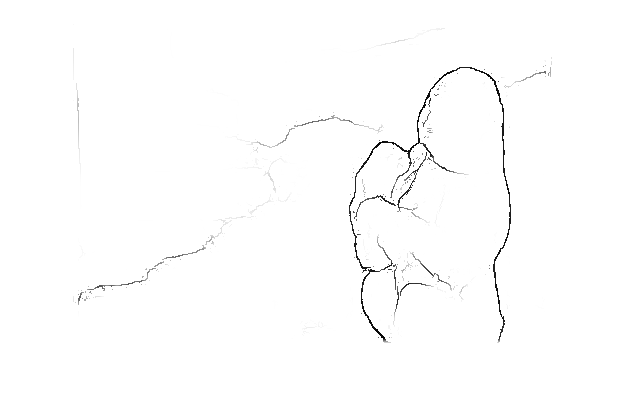}}
\subfloat[M-DSBD]{\includegraphics[width=0.49\linewidth]{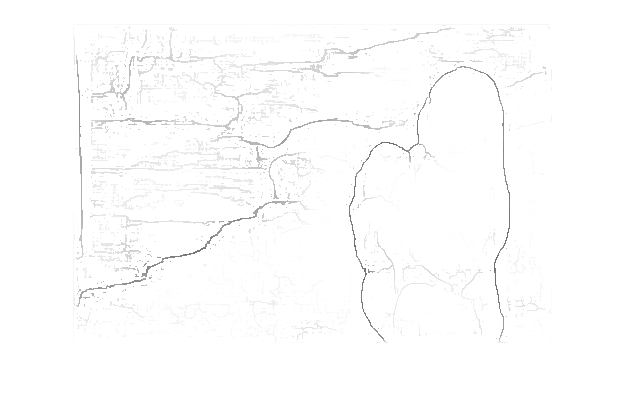}}
\vspace{-2.5ex}
\caption{(a) Shows an image from the BSDS500 dataset, (b) shows the groundtruth edge annotations (c) shows the edge output from HED and (d) shows the boundary output from M-DSBD.}
\label{fig:bsds}
\vspace{-5ex}
\end{figure}

\noindent\textbf{BSDS500:} For completeness, we report the performance of M-DSBD on the BSDS500 dataset. Table \ref{tab:bsds} tabulates the results. Note that M-DSBD was trained on the PASCAL Boundaries' \texttt{train} set, but tested on the BSDS500's \texttt{test} set. The numbers show that our model transfers to a different dataset while producing competitive results. Fig. \ref{fig:bsds} shows an example image from the BSDS500 dataset along with the edge and boundary detections. We can see from the figure M-DSBD transfers on to the BSDS500 dataset and is successful in providing high confidence for object boundaries and low confidence for internal edges.
\vspace{-1ex}
\begin{table}
\centering
\begin{tabular}{|c|c|c|c|}
\hline
Method & ODS & OIS & AP\\
\hline \hline
SE \cite{dollar2014fast} & 0.746 & 0.767 & 0.803 \\
HED \cite{xie2015holistically} & 0.782 & 0.804 & 0.833 \\
M-DSBD-PASCAL-B & 0.751 & 0.773 & 0.789 \\
\hline
\end{tabular}
\vspace{-2ex}
\caption{Results on the BSDS500 dataset. The M-DSBD model was trained on the PASCAL Boundaries dataset. The results show that methods trained on a boundary detection task perform fairly well on the edge detection task.}
\label{tab:bsds}
\vspace{-4ex}
\end{table}


\section{Conclusion and Future Work}
\label{sec:conclusion}
\vspace{-1ex}
In this paper, we pointed to the ambiguity in the definition of edge detection, and, defined a precise task, namely class-agnostic boundary detection. To facilitate progress in solving this problem, we release a large dataset of $\sim10$k images with labeled boundaries, which is 20 times bigger than the widely-used BSDS500 dataset, and without any ambiguity in the annotations. In addition, we proposed a novel multi-scale deep semantic boundary detector and showed that it performs well on the boundary detection task.


We now conclude the paper by pointing to various new research directions that can emerge out of this dataset. Firstly, since boundaries are complementary to pixel-level semantic labeling, it would be interesting to develop joint techniques that can exploit the advantages of each of these respective tasks. Secondly, state-of-the-art object proposal generators are based on edge grouping. It will be interesting to study the effect that instance-level semantic boundary predictions have on object proposals. And, finally, this dataset allows easy access to regions of occlusions because of the presence of occlusion cues (triple points). This dataset provides a good starting point to work on the hard task of occlusion-handling.
{\small
\bibliographystyle{ieee}
\bibliography{egbib}
}

\end{document}